\documentclass[preprint,12pt]{elsarticle}

\newcommand{\JournalName}{Information Sciences}

\usepackage{amssymb}
\usepackage{amsmath}
\usepackage{pifont}
\usepackage{natbib}
\usepackage{microtype}
\usepackage{multirow}
\usepackage{booktabs}
\usepackage[table,xcdraw]{xcolor}
\usepackage[normalem]{ulem}
\useunder{\uline}{\ul}{}

\usepackage{tabularx}
\usepackage{graphicx}
\usepackage{float}
\usepackage{caption}
\usepackage{subcaption}
\usepackage{hyperref}
\hypersetup{colorlinks=true, linkcolor=cyan, filecolor=magenta, urlcolor=cyan, citecolor=cyan}
\usepackage{cleveref}
\crefformat{figure}{#2Fig.~#1#3}
\crefformat{table}{#2Table~#1#3}
\crefformat{equation}{#2Eq.~#1#3}
\crefformat{section}{#2Section~#1#3}

\journal{\JournalName}

\begin{document}
\begin{frontmatter}
	\title{Robust Partial 3D Point Cloud Registration via Confidence Estimation under Global Context}

	\author[label1]{Yongqiang Wang}
	\author[label1]{Weigang Li\texorpdfstring{\corref{cor1}}{*}}
	\cortext[cor1]{Corresponding author}
	\ead{liweigang.luck@foxmail.com}
	\author[label2]{Wenping Liu}
	\author[label1]{Zhe Xu}
	\author[label1]{Zhiqiang Tian}

	\address[label1]{Engineering Research Center for Metallurgical Automation and Measurement Technology Ministry of Education, Wuhan University of Science and Technology, Wuhan 430081, China}
	\address[label2]{School of Information Management and Institute of Big Data and Digital Economy, Hubei University of Economics, Wuhan, 430205, China}

    \begin{abstract}
		Partial point cloud registration is essential for autonomous perception and 3D scene understanding, yet it remains challenging owing to structural ambiguity, partial visibility, and noise. We address these issues by proposing Confidence Estimation under Global Context (CEGC), a unified, confidence-driven framework for robust partial 3D registration. CEGC enables accurate alignment in complex scenes by jointly modeling overlap confidence and correspondence reliability within a shared global context. Specifically, the hybrid overlap confidence estimation module integrates semantic descriptors and geometric similarity to detect overlapping regions and suppress outliers early. The context-aware matching strategy smitigates ambiguity by employing global attention to assign soft confidence scores to correspondences, improving robustness. These scores guide a differentiable weighted singular value decomposition solver to compute precise transformations. This tightly coupled pipeline adaptively down-weights uncertain regions and emphasizes contextually reliable matches. Experiments on ModelNet40, ScanObjectNN, and 7Scenes 3D vision datasets demonstrate that CEGC outperforms state-of-the-art methods in accuracy, robustness, and generalization. Overall, CEGC offers an interpretable and scalable solution to partial point cloud registration under challenging conditions.
    \end{abstract}



    \end{frontmatter}


\section{INTRODUCTION}
\label{sec:Introduction}
3D point cloud registration is a fundamental problem in computer vision that facilitates the spatial alignment of geometric data captured from different viewpoints~\cite{lyu2024rigid}. It underpins a variety of downstream applications, such as 3D reconstruction, autonomous driving, simultaneous localization and mapping, and robotic perception~\cite{zhao2024deep, zhang2025random}. However, captured point clouds in real-world scenarios are often incomplete owing to occlusions, sensor limitations, or restricted viewpoints. Consequently, the assumption of full overlap between the source and target point clouds rarely holds, making partial-to-partial registration common and significantly more challenging.

Compared with fully overlapping registration, the partial setting requires accurate identification of overlapping regions, rejection of outliers from non-overlapping areas, and estimation of transformations under severe structural ambiguity~\cite{li2024dbdnet}. However, these requirements become challenging in cluttered or dynamic environments, where reliable alignment is essential for consistent scene understanding. Despite the increasing relevance of these challenges in real-world applications, robust and generalizable solutions for partial registration remain largely underexplored.

Prior approaches can be broadly categorized into classical geometric methods and modern learning-based frameworks. Classical techniques such as Iterative Closest Point (ICP) ~\cite{besl1992method} and Fast Global Registration (FGR) ~\cite{zhou2016fast} rely on nearest neighbor search and rigid transformation optimization. However, they assume strong initial alignment and high overlap, which limits their effectiveness in partial or noisy settings. By contrast, deep learning methods have demonstrated promising capabilities in learning discriminative features and robust correspondences to enhance performance in challenging settings~\cite{ma2022effective, hu2022daniel}. Nevertheless, most existing learning-based frameworks struggle to generalize under partial-to-partial conditions~\cite{jiang2025hybrid}. A key limitation lies in their limited ability to reason jointly on semantic and geometric structure, which hinders accurate overlap localization. Furthermore, most existing methods overlook the uncertainty inherent in correspondence matching, resulting in error propagation during transformation estimation~\cite{wu2025multi}. The absence of an integrated framework that jointly models region overlap, correspondence reliability, and transformation consistency under uncertainty remains a critical gap in the current literature.

To address these limitations, we propose Confidence Estimation under Global Context (CEGC), a unified confidence-driven framework for robust partial 3D point cloud registration. In particular, CEGC jointly models overlap confidence and correspondence reliability within a shared global context, enabling more accurate and robust alignment in complex scenes. We propose a Hybrid Overlap Confidence Estimation (HOCE) module that fuses semantic descriptors with geometric pairwise similarities to accurately identify overlapping regions and filter out non-overlapping points at an early stage. Moreover, we mitigate structural ambiguity by proposing a Context-Aware Matching Strategy (CAMS) that leverages attention-based global reasoning to infer soft confidence scores for point correspondences. Subsequently, these scores are used by a confidence-guided solver to compute the final transformation. This tightly coupled pipeline ensures that uncertain regions are preliminarily down-weighted, correspondences are contextually filtered, and the final alignment remains robust under challenging conditions, such as noise or partial visibility. Overall, CEGC offers a unified, interpretable solution to partial 3D point cloud registration and provides new insights to enhance robustness under incomplete observations and structural ambiguity.

\begin{figure}[!ht]
	\centering
	\includegraphics[width=0.9\linewidth,keepaspectratio]{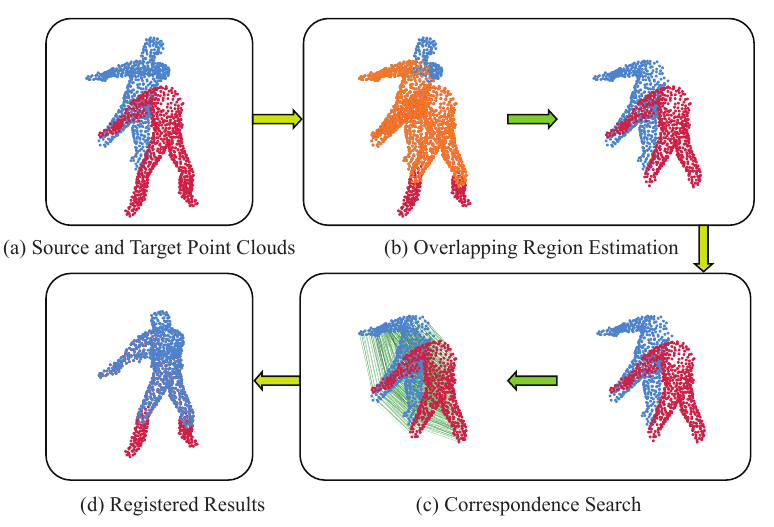}
	\caption{Overview of the proposed framework. (a) Input source (red) and target (blue) point clouds; (b) estimation of overlapping regions (orange); (c) correspondence inference, where the depth of green lines indicates the confidence of point-wise associations; (d) final registration result.}
	\label{fig:IntroductionOverview}
\end{figure}

The point cloud processing pipeline of the proposed method is illustrated in \cref{fig:IntroductionOverview}. Given a pair of partially overlapping point clouds (a), the HOCE module identifies reliable overlapping regions by jointly leveraging semantic and geometric cues (b). The CAMS then infers soft correspondences with confidence weights based on global contextual reasoning (c). Finally, a confidence-guided solver aggregates these weighted correspondences to estimate the rigid transformation that aligns the input point clouds (d).

The main contributions of this work are summarized as follows:
\begin{itemize}
    \item We propose an HOCE module that integrates semantic and geometric cues to identify overlapping areas and suppress outliers preliminarily.

    \item We design a CAMS strategy that leverages global attention to assign soft confidence to correspondences, enhancing robustness to structural ambiguity and noise.

    \item We validate our proposed method across diverse datasets (ModelNet40, ScanObjectNN, and 7Scenes), achieving state-of-the-art performance in accuracy, robustness, and generalization without relying on dataset-specific tuning.
\end{itemize}

The remainder of this work is organized as follows.
\cref{sec:RELATED WORK} reviews the landscape of existing methods, including geometric, learning-based, and partial registration approaches.
\cref{sec:PROBLEM FORMULATION} formulates the partial registration problem and presents the overall design goals.
\cref{sec:METHOD} elaborates on the proposed CEGC framework, detailing the architecture and the role of each component.
\cref{sec:EXPERIMENTS AND DISCUSSIONS} presents comprehensive experiments across synthetic and real-world datasets, with comparative and ablation results that demonstrate the effectiveness of our method.
Finally, \cref{sec:CONCLUSION} concludes the paper and discusses potential directions for future research.

\section{RELATED WORK}
\label{sec:RELATED WORK}
In computer vision and robotics, 3D point cloud registration is a core task that is essential for applications such as robotic perception, 3D reconstruction, autonomous navigation, and object-level understanding~\cite{lyu2024rigid, ma2022effective}. The registration problem becomes more challenging under partial visibility, where point clouds captured from different viewpoints exhibit limited overlap owing to occlusion, restricted field of view, or sensor sparsity~\cite{wu2024surface}. Existing registration methods are broadly categorized into classical geometric pipelines, deep learning-based frameworks, and recent work addressing partial registration with varying degrees of confidence modeling.

\subsection{Classical Geometric Registration}
Traditional point cloud registration methods generally rely on iterative geometric optimization by assuming high overlap and low noise~\cite{cao2024new, yan2024discriminative}. The most representative approach is ICP~\cite{besl1992method}, which alternates between nearest neighbor correspondence search and closed-form singular value decomposition for transformation estimation. Despite its simplicity and efficiency, ICP is highly sensitive to initialization and often converges to local minima, particularly in low-overlap or cluttered scenes.

Various enhancements have been proposed to improve robustness. These methods include Go-ICP~\cite{yang2015go}, which integrates a branch-and-bound global search, and RANSAC-based techniques~\cite{huang2024efficient} that use random sampling and voting to reject outliers. However, these approaches also assume high overlap and suffer from high computational costs or degeneracy under large outlier ratios~\cite{kolpakov2023approach}. Therefore, their applicability is limited in real-world, partial-to-partial registration settings.

\subsection{Learning-Based Registration}
With the advancement of geometric deep learning, numerous data-driven registration methods that aim to learn feature representations and transformation solvers in an end-to-end manner have emerged. Deep Closest Point (DCP)~\cite{wang2019deep} combines DGCNN-based local feature extraction with Transformer-style global reasoning to generate soft correspondences, which is followed by transformation estimation using SVD. In addition, a series of subsequent models, such as PointDifformer~\cite{she2024pointdifformer} and PARE-Net~\cite{yao2024pare}, further extend this paradigm by exploring alternative strategies including global feature regression and similarity matrix learning.

Building on these early explorations, subsequent works have further advanced registration by incorporating richer structural cues. For example, PRNet~\cite{wang2019prnet} and GeoTransformer~\cite{qin2023geotransformer} integrate local and global geometric information through attention and Transformer-style architectures. Complementary to these designs, FMR~\cite{huang2020feature} proposes a feature-metric formulation that aligns point clouds within a learned similarity space, while methods such as REGTR~\cite{yew2022regtr} emphasize coarse-to-fine alignment and relational correspondence modeling.

However, existing learning-based methods either assume full overlap or use fixed thresholds to handle outliers, limiting their robustness under partial visibility. Furthermore, while attention enhances structural reasoning, most frameworks do not explicitly quantify or propagate correspondence uncertainty across the pipeline.

\subsection{Toward Confidence-Aware Partial Registration}
Partial-to-partial registration requires feature matching and confidence modeling to suppress unreliable regions and outliers. Existing methods differ in their confidence target, representation, context, and coupling to pose estimation.

Overlap-based: OMNet~\cite{xu2021omnet} predict region-level overlap. While effective against clutter, they remain coarse and local, lacking fine-grained correspondence reliability.

Correspondence-based: RPMNet~\cite{yew2020rpm} produces soft correspondences via Sinkhorn (implicit confidence). FINet~\cite{xu2022finet} and DBDNet~\cite{li2024dbdnet} make match reliability explicit but treat overlap, matching, and pose as modular stages, limiting consistent uncertainty propagation.

Global but confidence-agnostic: REGTR~\cite{yew2022regtr} uses global Transformers to directly predict correspondences but without explicit confidence, reducing robustness in ambiguous settings.

Unlike extant research, CEGC disentangles overlap and correspondence confidence, reasons jointly under a global context, and couples overlap confidence and correspondence confidence to a confidence-guided solver. This yields fine-grained, globally consistent, and solver-aligned confidence, crucial for robustness under low overlap and noise.

\begin{table}[htbp]
\centering
\caption{Confidence modeling in representative registration methods.}
\label{tab:confidence_comparison}
\scriptsize
\begin{tabularx}{\linewidth}{l *{4}{>{\centering\arraybackslash}X}}
\toprule
Method & Target & Representation & Context & Coupling \\
\midrule
OMNet~\cite{xu2021omnet} & Overlap & Hard mask & Local & Decoupled \\
RPMNet~\cite{yew2020rpm} & Correspondence & Implicit & Local & Decoupled \\
FINet~\cite{xu2022finet} & Correspondence & Explicit score & Local & Stage-wise \\
DBDNet~\cite{li2024dbdnet} & Correspondence & Refined score & Local & Stage-wise \\
REGTR~\cite{yew2022regtr} & --- & None & Global & End-to-end \\
CEGC (ours) & Overlap \& Corr. & \mbox{Explicit, disentangled} & Global & End-to-end \\ \bottomrule
\end{tabularx}
\end{table}

Beyond the field of point cloud registration, recent progress in active learning and learnability~\cite{zhang2024learnability} has exploited model uncertainty in broader machine learning contexts. These insights are conceptually related to our confidence estimation strategy, providing additional theoretical support for designing robust and generalizable registration frameworks.

\section{PROBLEM FORMULATION}
\label{sec:PROBLEM FORMULATION}
\subsection{Notation}
\label{sec:Notation}
Let the source point cloud be denoted by $\mathbf{X} = \{x_1, x_2, \ldots, x_{N_X}\} \in \mathbb{R}^{N_X \times 3}$ and the target point cloud by $\mathbf{Y} = \{y_1, y_2, \ldots, y_{N_Y}\} \in \mathbb{R}^{N_Y \times 3}$, where $N_X$ and $N_Y$ represent the number of points in $\mathbf{X}$ and $\mathbf{Y}$, respectively. In partial-to-partial registration, both $\mathbf{X}$ and $\mathbf{Y}$ represent incomplete observations of a shared underlying 3D structure.

A rigid transformation is parameterized by a rotation matrix $\mathbf{R} \in SO(3)$ and a translation vector $\mathbf{t} \in \mathbb{R}^3$, jointly forming $\mathbf{T} = (\mathbf{R}, \mathbf{t})$. Applying this transformation to a point $x \in \mathbf{X}$ is expressed as $\mathbf{T}(x) = \mathbf{R}x + \mathbf{t}$. The estimated transformation is denoted as $\hat{\mathbf{T}} = (\hat{\mathbf{R}}, \hat{\mathbf{t}})$, while the ground truth is represented by $\mathbf{T}^* = (\mathbf{R}^*, \mathbf{t}^*)$. We denote the transformed source set by $\hat{\mathbf{X}} = \{\hat{x}_i = \hat{\mathbf{T}}(x_i)\}_{i=1}^{N_X}$.

We use $\cdot$ to denote the inner product, and $\times$ for the vector cross product or scalar multiplication, depending on context. The notation $\| \cdot \|^2$ represents the squared Euclidean (or Frobenius) norm, while $|\cdot|$ denotes the absolute value of a scalar.

\subsection{Problem Definition}
\label{sec:Problem Definition}
Given two partially overlapping point clouds $\mathbf{X}$ and $\mathbf{Y}$, the goal of partial point cloud registration is to estimate the optimal transformation $\hat{\mathbf{T}} = (\hat{\mathbf{R}}, \hat{\mathbf{t}})$ such that the transformed source $\hat{\mathbf{X}}$ aligns with the target $\mathbf{Y}$. Unlike full-to-full registration, where all points have potential correspondences, the partial setting introduces unknown and potentially large non-overlapping regions, significantly increasing the difficulty of establishing reliable point-wise correspondences.

Formally, the registration problem is cast as the following optimization:
\begin{equation}
	\hat{\mathbf{T}} = \underset{\mathbf{R} \in SO(3),\ \mathbf{t} \in \mathbb{R}^3}{\arg\max}\ \mathcal{L}(\mathbf{R}, \mathbf{t}; \mathbf{X}, \mathbf{Y}),
\end{equation}
where $\mathcal{L}(\cdot)$ represents a loss function that quantifies the misalignment between the transformed source $\mathbf{R}{x}_i + \mathbf{t}$ and the target ${Y}$, often involving a correspondence term, a rejection strategy for outliers, and potentially geometric or semantic consistency priors.

In the absence of known correspondences, this becomes a joint optimization over transformation and correspondence estimation, often requiring robust initialization or learning-based models. The key challenge lies in identifying the overlapping regions between $\mathbf{X}$ and $\mathbf{Y}$, while avoiding being misled by outliers and non-overlapping structures.

\section{METHOD}
\label{sec:METHOD}
Partial-to-partial point cloud registration presents significant challenges owing to overlap uncertainty, ambiguous correspondences, and transformation estimation errors in the presence of outliers. We tackle these challenges systematically using an end-to-end registration framework named CEGC. CEGC integrates semantic-geometric reasoning and confidence-aware correspondence matching within a unified architecture.

\begin{figure}[!htbp]
	\centering
	\includegraphics[width=\linewidth,keepaspectratio]{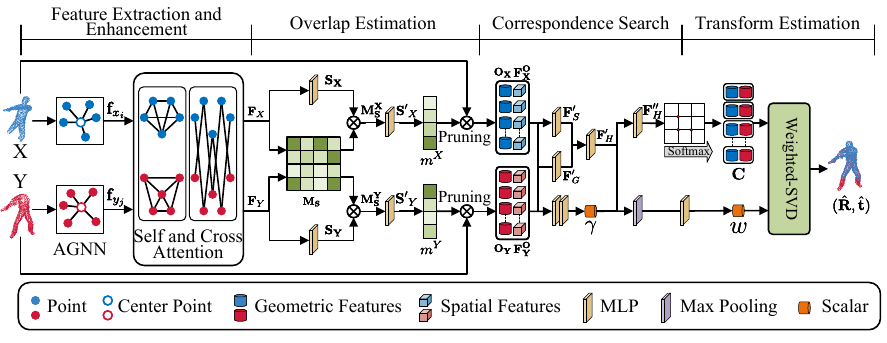}
	\caption{Overview of the proposed registration framework.
	The pipeline entails four stages: (a) local feature extraction and attention-based enhancement; (b) overlap prediction using a hybrid confidence estimator that integrates semantic and geometric cues; (c) context-aware correspondence matching with attention-based refinement; and (d) transformation estimation via confidence-weighted SVD.}
	\label{fig:Fig_Overview}
\end{figure}

As illustrated in \cref{fig:Fig_Overview}, the proposed framework comprises four key stages. First, local geometric features are extracted using an attention-based graph neural network (AGNN), augmented with self- and cross-attention to capture both intra- and inter-cloud context. Second, the HOCE module predicts point-wise overlap likelihood by fusing semantic and geometric cues, enabling early removal of non-overlapping regions. Third, CAMS generates dense correspondence candidates and refines them via global attention, producing soft confidence scores to handle ambiguity and noise. Finally, a confidence-weighted SVD solver estimates the rigid transformation. The entire pipeline is jointly trained in an end-to-end manner for robust, accurate, and generalizable registration in partial-view scenarios.

\subsection{Feature Extraction and Enhancement}
To support robust overlap prediction, we first extract expressive geometric features and enhance them with global contextual cues. This process includes two key stages: local feature construction using a Graph Neural Network (GNN) and context enrichment via self- and cross-attention.

\subsubsection{Local Feature Construction}
\begin{figure}[!htbp]
	\centering
	\includegraphics[width=0.8\linewidth,keepaspectratio]{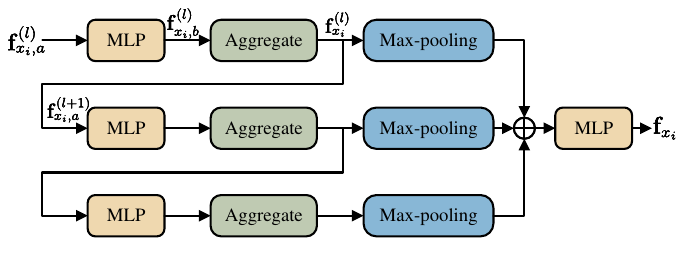}\\
	\caption{Architecture of AGNN.}
	\label{fig:FeatureExtraction}
\end{figure}

Graph learning has demonstrated strong potential in 3D vision. For instance, HOGFormer~\cite{xie2025hogformer} integrates high-order graph convolutions with Transformers for human pose estimation and Stacked Capsule Graph Autoencoders~\cite{hong2021stacked} leverage geometry-aware capsule graphs for head pose estimation. Inspired by these advances, we introduce an Adaptive GNN (AGNN) tailored for partial point cloud registration, enabling adaptive local-global reasoning for robust confidence modeling~\cite{tan2023using}. The AGNN extracts informative local descriptors by constructing neighborhood-aware feature representations, as illustrated in \cref{fig:FeatureExtraction}.

Each point $x_i \in \mathbf{X}$ gathers information from its $K$-nearest neighbors $\mathcal{N}(x_i)$, with their relative positions encoded as
\begin{equation}
\mathbf{H}_{x_i} = \left\{ [x_i,\ x_i - x_i^j] \;\middle|\; x_i^j \in \mathcal{N}(x_i) \right\} \in \mathbb{R}^{K \times 6}.
\end{equation}

An attention-based aggregation updates the point-wise feature at each layer:
\begin{align}
\mathbf{f}_{x_i}^{(l)} &= f_{C_1} \left( \sum_{j=1}^{K} a_{ij}^{(l)} \cdot \mathbf{f}_{x_j, b}^{(l)} \cdot \mathbf{W}^V \right) + \mathbf{f}_{x_i, b}^{(l)}~, \\
a_{ij}^{(l)} &= \text{softmax} \left( \frac{(\mathbf{f}_{x_i}^{(l)} \mathbf{W}^Q)(\mathbf{f}_{x_j}^{(l)} \mathbf{W}^K)^\top}{\sqrt{d_l}} \right),
\end{align}
where $f_{C_1}$	represents a multi-layer perceptron (MLP) that has an output dimensionality of $C_1$, $a_{ij}^{(l)} \in \mathbb{R}$ represents the attention weight between the center point $x_i$ and its neighbor $x_j$ at layer $l$, and $\mathbf{W}^Q$, $\mathbf{W}^K$, and $\mathbf{W}^V$ represent learnable projection matrices. Here, $\mathbf{f}^{(1)}_{x_i, a} = \mathbf{H}_{x_i}$, $\mathbf{f}^{(l)}_{x_i, b} = \text{MLP}(\mathbf{f}^{(l)}_{x_i, a})$, and $\mathbf{f}^{(l+1)}_{\mathbf{x}_i, a} = \mathbf{f}^{(l)}_{\mathbf{x}_i}$.

Final features are obtained via max-pooling across multiple layers:
\begin{equation}
\mathbf{f}_{x_i} = f_{C_2} \left( \left[ \zeta(\mathbf{f}_{x_i}^{(1)}),\ \zeta(\mathbf{f}_{x_i}^{(2)}),\ \zeta(\mathbf{f}_{x_i}^{(3)}) \right] \right).
\end{equation}
The same process is applied to the target point cloud $\mathbf{Y}$.

\subsubsection{Context Enhancement via Self- and Cross-Attention}
To promote semantic consistency between point clouds, we utilize interleaved self-attention and cross-attention layers. These attention mechanisms enable each point to adaptively incorporate information from both its source and counterpart clouds, ultimately enhancing feature representations for precise overlapping region estimation.

\textbf{Self-Attention (within $\mathbf{X}$ or $\mathbf{Y}$):} Within each point cloud, features are refined by attending to all other points, which is mathematically expressed as follows:
\begin{equation}
	\mathbf{f}_{x_i} \leftarrow \mathbf{f}_{x_i} + f_{C_2} \left( \left[ \mathbf{f}_{x_i},\ \sum_j e_{ij} \mathbf{v}_{x_j} \right] \right),
\end{equation}
\begin{equation}
	e_{ij} = \text{softmax} \left( \frac{ \mathbf{q}^{x_i} (\mathbf{k}^{x_j})^\top }{ \sqrt{d} } \right),
\end{equation}
where $\mathbf{q}_{x_i}$, $\mathbf{k}_{x_j}$, and $\mathbf{v}_{x_j}$ represent the query, key and value vectors, respectively, derived from the same point cloud.

\textbf{Cross-Attention (from $\mathbf{X} \rightarrow \mathbf{Y}$ or vice versa):} To encode cross-cloud context, features of $\mathbf{X}$ are updated using points from $\mathbf{Y}$:
\begin{equation}
	\mathbf{f}_{x_i} \leftarrow \mathbf{f}_{x_i} + f_{C_2} \left( \left[ \mathbf{f}_{x_i},\ \sum_j e_{ij} \mathbf{v}_{y_j} \right] \right),
\end{equation}
\begin{equation}
	e_{ij} = \text{softmax} \left( \frac{ \mathbf{q}_{x_i} (\mathbf{k}_{y_j})^\top }{ \sqrt{d} } \right),
\end{equation}
where $\mathbf{q}_{x_i}$, $\mathbf{k}_{y_j}$, and $\mathbf{v}_{y_j}$ represent the query, key, and value vectors, respectively, derived from the different point clouds.

Following alternation, we obtain the enhanced features as
\begin{equation}
	\mathbf{F}_{X} = \left\{ \mathbf{f}_{x_i} \right\}_{i=1}^{N_X} \in \mathbb{R}^{N_X \times {C_2}}, \quad
	\mathbf{F}_{Y} = \left\{ \mathbf{f}_{y_j} \right\}_{j=1}^{N_Y} \in \mathbb{R}^{N_Y \times {C_2}},
\end{equation}
which encode both local geometry and global correspondence cues.

\subsection{Overlap Estimation via HOCE}
Accurately identifying overlapping regions is essential for partial-to-partial registration. To this end, we propose an HOCE model that integrates two complementary branches: semantic-based confidence and cross-geometric similarity. This hybrid design enhances the reliability of overlap prediction while mitigating noise introduced by non-overlapping structures.

\textbf{Semantic Confidence Branch:} This branch estimates the per-point overlap confidence for $\mathbf{X}$ based solely on its extracted features, without directly interacting with $\mathbf{Y}$:
\begin{equation}
	\mathbf{S}_\mathbf{X} = \sigma(f(\mathbf{F}_{\mathbf{X}})) \in \mathbb{R}^{N_X \times 1},
\end{equation}
where $f$ represents MLP, whose output dimension is 1 and $\sigma$ represents the sigmoid activation.

\textbf{Geometric Similarity Branch:} This branch computes the cosine similarity between point pairs $(x_i, y_j)$ from $\mathbf{X}$ and $\mathbf{Y}$ to capture geometric correspondences:
\begin{equation}
	\mathcal{W}_{ij} = \frac{\mathbf{f}_{x_i} \cdot \mathbf{f}_{y_j}}{\lvert \mathbf{f}_{x_i} \rvert \times \lvert \mathbf{f}_{y_j} \rvert}, \quad \forall i \in [1, N_X],\ j \in [1, N_Y],
\end{equation}
\begin{equation}
	\mathbf{M}_{\mathbf{S}} = [\mathcal{W}_{ij}] \in \mathbb{R}^{N_X \times N_Y}.
\end{equation}

\textbf{Hybrid Confidence Fusion:} The semantic and geometric cues are fused via elementwise multiplication:
\begin{equation}
	\mathbf{M}_{\mathbf{S}}^{\mathbf{X}} = \mathbf{M}_{\mathbf{S}} \otimes \mathbf{S}_{X},
\end{equation}
where $\otimes$ represents row-wise broadcasting and elementwise multiplication. Subsequently, this fused representation is refined by another MLP ($f$) followed by a sigmoid activation to produce the final overlap confidence:
\begin{equation}
	\mathbf{S'}_{X} = \sigma(f(\mathbf{M}_{\mathbf{S}}^{\mathbf{X}})).
\end{equation}

We then extract the top-$N$ most confident points:
\begin{equation}
	\mathcal{I} = \underset{\text{top-}N}{\arg\max}(\mathbf{S'}_{X}),
\end{equation}
and define a binary mask $m_i^X$ accordingly:
\begin{equation}
	m_i^{X} =
	\begin{cases}
		1, & \text{if } i \in \mathcal{I} \\
		0, & \text{otherwise}
	\end{cases}~, \quad \forall i \in [1, N_X].
	\end{equation}

The final overlapping region in $\mathbf{X}$ is extracted by masking:
\begin{equation}
	\mathbf{O}_{\mathbf{X}} = m^{X} \otimes \mathbf{X}~, \quad \mathbf{O}_{\mathbf{X}} \in \mathbb{R}^{N \times 3}.
\end{equation}

The same steps are applied to $\mathbf{Y}$ to obtain $\mathbf{O}_{\mathbf{Y}}$. These subsets $\mathbf{O}_{\mathbf{X}}$, $\mathbf{O}_{\mathbf{Y}}$ and their corresponding features $\mathbf{F}_{\mathbf{X}}^{\mathbf{O}}$, $\mathbf{F}_{\mathbf{Y}}^{\mathbf{O}}$ constitute the basis for the subsequent correspondence matching module.

\subsection{Correspondence Search via CAMS}
We propose the CAMS module to establish robust correspondences between overlapping regions identified by HOCE. The CAMS module integrates both local similarity cues and global structural context. CAMS first constructs pairwise descriptors that encode geometric relations between candidate point pairs. However, relying solely on local features can result in ambiguous or structurally inconsistent matches. We address this by introducing a lightweight yet effective refinement module—Context Feature Modulation (CFM)—which incorporates a global summary vector aggregated from all candidate correspondences. This global context serves as a structural prior to modulate local match confidence, encouraging mutual compatibility across the correspondence space. CAMS achieves more coherent and reliable correspondence estimation by coupling fine-grained similarity encoding with global structural reasoning.

\subsubsection{Pairwise Contextual Feature Construction}
Given the extracted overlapping subsets $\mathbf{O}_{\mathbf{X}} \in \mathbb{R}^{N \times 3}$ and $\mathbf{O}_{\mathbf{Y}} \in \mathbb{R}^{N \times 3}$, along with their associated features $\mathbf{F}_{\mathbf{X}}^{\mathbf{O}} \in \mathbb{R}^{N \times {C_2}}$ and $\mathbf{F}_{\mathbf{Y}}^{\mathbf{O}} \in \mathbb{R}^{N \times {C_2}}$, we first construct dense pairwise features for all candidate correspondences.

Each candidate pair $(o_{x_i}, o_{y_j})$ is encoded using spatial and feature-level information. The spatial descriptor is defined as
\begin{equation}
	\mathbf{F}_{S}(i, j) = \left[o_{x_i}, o_{y_j}, \| o_{x_i} - o_{y_j} \|\right], \quad \mathbf{F}_{S} \in \mathbb{R}^{N \times N \times C_3}.
\end{equation}

The geometric descriptor based on learned features is
\begin{equation}
	\mathbf{F}_{G}(i, j) = \left[\mathbf{f}_{x_i}^{o}, \ \mathbf{f}_{y_j}^{o}, \| \mathbf{f}_{x_i}^{o} - \mathbf{f}_{y_j}^{o} \|\right], \quad \mathbf{F}_{G} \in \mathbb{R}^{N \times N \times C_4}.
\end{equation}

Both descriptors are independently processed through MLPs, which is expressed as
\begin{equation}
	\mathbf{F}'_{S} = f_{C_5}(\mathbf{F}_{S})\quad \text{and} \quad \mathbf{F}'_{G} = f_{C_5}(\mathbf{F}_{G}).
\end{equation}

Subsequently, these representations are fused via elementwise interaction and further processed to produce the final pairwise embedding $\mathbf{F}'_{H}$:
\begin{equation}
	\mathbf{F}'_{H} = f_{C_5}(\mathbf{F}'_{S} \, \mathbf{F}'_{G}), \quad \mathbf{F}'_{H} \in \mathbb{R}^{N \times N \times C_5}.
\end{equation}

This embedding serves as the initial confidence map for match estimation, capturing local geometric similarity and feature alignment.

\subsubsection{Context Feature Modulation}
Given the pairwise descriptor $\mathbf{F}'_H$ constructed from local geometric relations, we aim to enhance its reliability by incorporating global structural context. While $\mathbf{F}'_H$ effectively encodes local similarity cues, it lacks sensitivity to higher-order dependencies across the full correspondence space. Consequently, matches that appear locally plausible may still conflict with the broader geometric structure. We resolve this limitation using the CFM refinement module to inject global context into the confidence estimation process. The CFM modulates individual matching scores in a globally consistent manner by aggregating a compact summary vector from all candidate correspondences, reinforcing coherence across the match set.

For each candidate correspondence $(o_{x_i}, o_{y_j})$, we begin by embedding its geometric relationship into a latent representation:
\begin{equation}
\mathbf{F}_{ij}^o = f_{C_6} \left( \left[ o_{x_i},\ o_{y_j},\ \|o_{x_i} - o_{y_j}\| \right] \right), \quad \mathbf{F}_{ij}^o \in \mathbb{R}^{N \times N \times C_6}.
\end{equation}

We incorporate global structure awareness by computing a global summary feature, which involves performing average pooling over all pairwise descriptors:
\begin{equation}
\mathbf{m} = \frac{1}{N^2} \sum_{i,j} \mathbf{F}_{ij}^o~.
\end{equation}

This compact global feature vector $\mathbf{m}$ serves as a shared context that guides subsequent match refinement. In particular, we compute a modulation weight for each candidate pair using a shared function, which is expressed as
\begin{equation}
\gamma(i,j) = \sigma \left( f_{\text{mod}} \left( \mathbf{F}_{ij}^o,\ \mathbf{m} \right) \right),
\end{equation}
where $f_{\text{mod}}$ represents a lightweight MLP applied channel-wise and $\sigma(\cdot)$ represents the sigmoid activation. This modulation mechanism enforces consistency between each local pair and the global structural context.

Finally, the modulated confidence map is computed by reweighting the initial descriptor $\mathbf{F}'_H$:
\begin{equation}
\mathbf{F}''_{H} = f_{C_7} \left( \gamma \cdot \mathbf{F}'_H \right),
\end{equation}
where $f_{C_7}$ represents a learnable refinement layer that outputs the final correspondence scores.

\begin{figure}[!htbp]
\centering
\includegraphics[width=0.9\linewidth,keepaspectratio]{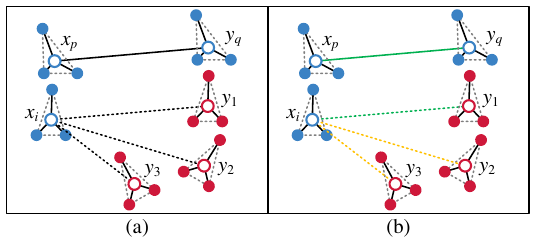}
\caption{Comparison of correspondence matching with and without context feature modulation.
(a) Without CFM, ambiguous matches (e.g., $x_i$ to $y_1$, $y_2$, $y_3$) arise due to reliance on local similarity alone.
(b) With CFM, confident correspondences such as $(x_p, y_q)$ act as contextual anchors that refine other matches based on global structure.}
\label{fig:fig_CorrespondingSearch}
\end{figure}

The effectiveness of the CFM module is qualitatively illustrated in \cref{fig:fig_CorrespondingSearch}. As shown in \cref{fig:fig_CorrespondingSearch}(a), relying solely on local similarity can result in ambiguous associations. For instance, $x_i$ may be linked to multiple uncertain candidates, such as $y_1$, $y_2$, or $y_3$. By contrast, \cref{fig:fig_CorrespondingSearch}(b) demonstrates that globally consistent correspondences, such as $(x_p, y_q)$, provide reliable structural cues that suppress implausible matches (green dotted line) and strengthen coherent ones (yellow dotted line). CFM modulates local confidence scores based on global structure to enhance both the robustness and consistency of correspondence estimation.

\subsection{Confidence-Guided Transformation Estimation}
Given the confidence-weighted correspondence matrix $\mathbf{F}''_{H}$ produced by CAMS, we estimate the rigid transformation that aligns the source point cloud $\mathbf{X}$ with the target point cloud $\mathbf{Y}$. The primary objective is to recover the optimal rotation $\hat{\mathbf{R}} \in \text{SO}(3)$ and translation $\hat{\mathbf{t}} \in \mathbb{R}^3$ that best align the two clouds based on the most reliable matches.

For each point $o_{x_i} \in \mathbf{O}_{\mathbf{X}}$, we select the corresponding point $o_{y_j} \in \mathbf{O}_{\mathbf{Y}}$ with the highest matching confidence:
\begin{equation}
	m(o_{x_i}) = \arg\max_j\ \mathbf{F}''_{H}(i, j).
\end{equation}

This results in a set of soft one-to-one correspondences:
\begin{equation}
	\mathbf{C} = \left\lbrace \left(o_{x_i}, m(o_{x_i})\right) \right\rbrace_{i=1}^{N},
\end{equation}
where $N$ indicates the total number of overlapping points identified between the source and target point sets. The term ``soft'' reflects the use of confidence scores to infer correspondences without relying on deterministic or ground-truth matches.

We formulate a confidence-weighted least squares optimization to estimate the transformation:
\begin{equation}
	(\hat{\mathbf{R}}, \hat{\mathbf{t}}) = \arg\min_{\mathbf{R}, \mathbf{t}} \sum_{i=1}^{M} w_i \left\| \mathbf{R} o_{x_i} + \mathbf{t} - m(o_{x_i}) \right\|^2,
\end{equation}
where $w_i \in [0, 1]$ represents the confidence weight associated with the $i$-th match. These weights modulate the influence of each correspondence on the final alignment.

\begin{figure}[!htbp]
	\centering
	\includegraphics[width=0.7\linewidth,keepaspectratio]{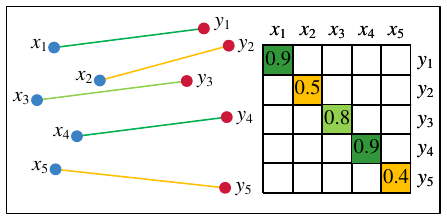}
	\caption{Visualization of the confidence-aware correspondence matrix $\mathbf{F}''_H$ used for transformation estimation. High-confidence matches such as $(x_1, y_1)$ and $(x_4, y_4)$ are highlighted in blue with scores close to 0.9, indicating strong semantic and geometric consistency. Lower-confidence matches (e.g., $(x_2, y_2)$, $(x_5, y_5)$) are downweighted to reduce the impact of noisy or ambiguous associations. These weights are used to guide the confidence-weighted SVD-based estimation of the rigid transformation.}
	\label{fig:Fig_CorrespondingWeighting}
\end{figure}

To visualize the effect of this weighting strategy, \cref{fig:Fig_CorrespondingWeighting} presents an example of the final confidence-aware correspondence matrix $\mathbf{F}''_H$. Each entry reflects the model's belief in the reliability of the corresponding match, taking into account both local similarity and global structural support via CAMW. For instance, $(x_1, y_1)$ and $(x_4, y_4)$ exhibit high scores ($\approx 0.9$) due to strong contextual consistency, while $(x_2, y_2)$ and $(x_5, y_5)$ are assigned lower weights ($\approx 0.4-0.5$), reflecting their weaker support or ambiguity. These weights directly impact the downstream transformation estimation process by emphasizing more trustworthy correspondences.

The confidence weights $w_i$ are derived from the product of the initial similarity scores $\mathbf{F}'_{H}$ and the context-aware confidence map $\gamma$ from CAMW. Specifically, for each point $o_{x_i}$, we compute:
\begin{equation}
\label{eq:SVD}
	w_i = \sigma \left( f \left( \max_i \left( \gamma \, \mathbf{F}'_{H} \right) \right) \right),
\end{equation}
where $f$ represents a learnable projection and $\sigma$ represents the sigmoid activation.

Following~\cite{chen2023full, yuan2024inlier}, we solve this optimization via the closed-form SVD approach. The weighted centroids of the matched point sets are computed as:
\begin{equation}
	\bar{o_x} = \frac{\sum_i w_i o_{x_i}}{\sum_i w_i} \quad \text{and} \quad \bar{o_y} = \frac{\sum_i w_i m(o_{x_i})}{\sum_i w_i}.
\end{equation}

We then construct the weighted covariance matrix:
\begin{equation}
	\mathbf{H} = \sum_{i=1}^{N} w_i (o_{x_i} - \bar{o_x})(m(o_{x_i}) - \bar{o_y})^\top,
\end{equation}
and perform SVD:
\begin{equation}
	\mathbf{H} = \mathbf{U} \Sigma \mathbf{V}^\top.
\end{equation}

Finally, The optimal rigid transformation is recovered as:
\begin{equation}
	\hat{\mathbf{R}} = \mathbf{V} \mathbf{U}^\top \quad \text{and} \quad  \hat{\mathbf{t}} = \bar{o_y} - \hat{\mathbf{R}} \bar{o_x}~.
\end{equation}

\subsection{Loss Function}
To jointly optimize overlap identification and transformation estimation, we adopt a multi-task loss that combines classification (for overlap estimation) and regression (for pose estimation) objectives.

Overlap estimation is formulated as a binary classification task, where each point is labeled as either overlapping or non-overlapping. Accordingly, a binary cross-entropy loss is applied independently to the source and target point clouds. For the source point clouds, the loss is defined as

\begin{equation}
	\mathcal{L}_1 = \frac{1}{2N_X} \sum_{i=1}^{N_X} \left[ m_i^x \log(\hat{m}_i^x) + (1 - {m}_i^x) \log(1 - \hat{m}_i^x) \right].
\end{equation}

Similarly, for the target point cloud:
\begin{equation}
	\mathcal{L}_2 = \frac{1}{2N_Y} \sum_{j=1}^{N_Y} \left[ {m}_j^y \log(\hat{m}_j^y) + (1 - {m}_j^y) \log(1 - \hat{m}_j^y) \right],
\end{equation}
where $\hat{m}_i^x$ and $\hat{m}_j^y$ represent the predicted overlap probabilities, and $m_i^x$, $m_j^y$ represent the corresponding ground-truth binary masks.

The total overlap estimation loss is given by
\begin{equation}
	\mathcal{L}_O = \mathcal{L}_1 + \mathcal{L}_2~.
\end{equation}

To supervise the estimated rigid transformation, we define a registration loss that penalizes the deviation between predicted and ground-truth pose. It includes both rotational and translational terms:
\begin{equation}
	\mathcal{L}_R = \| {R^*}^T \hat{R} - I \|^2 + \| t^* - \hat{t} \|^2~,
\end{equation}
where $\hat{R}, \hat{t}$ represent the estimated rotation and translation, $R^*, t^*$ represent the ground-truth transformations, and $I$ represent the identity matrix to enforce orthogonality in rotation.

The final objective function is a weighted combination of the two losses:
\begin{equation}
\mathcal{L}_{\text{all}} = \lambda \cdot \mathcal{L}_O + (1 - \lambda) \cdot \mathcal{L}_R~,
\end{equation}
where $\lambda \in [0, 1]$ balances the contributions of overlap prediction and pose estimation during training.

\section{EXPERIMENTS AND DISCUSSIONS}
\label{sec:EXPERIMENTS AND DISCUSSIONS}
This section presents a comprehensive evaluation of the proposed method. Comparative experiments are first conducted on the ModelNet40 dataset under partial overlap conditions. The evaluation covers generalization to unseen examples and object categories, as well as robustness under Gaussian noise. To assess applicability in real-world scenarios, cross-dataset experiments are carried out on two challenging benchmarks: ScanObjectNN, a real-world object dataset, and 7Scenes, a cluttered indoor scene dataset. Finally, ablation studies are performed to quantify the contribution of each module within the overall framework.

\subsection{Dataset}
\label{sec:Dataset}
The ModelNet40 benchmark~\cite{modelnet40} contains CAD models from 40 man-made object categories. It is a widely used dataset in the point cloud research community. For each shape, we uniformly sample 1,024 points from the raw mesh to construct the point cloud. The source and target point clouds are generated separately, with the target obtained by applying a rigid transformation to the source. In particular, three Euler angles are sampled independently within $[-45^\circ, 45^\circ]$ and translations along each axis from $[-1, 1]$, thereby simulating realistic spatial variations during both training and testing. We adopt the random cropping strategy of RPMNet~\cite{yew2020rpm} to generate partially overlapping point clouds. A random direction is sampled from the unit sphere and a cropping plane is defined through the centroid. Points within a specified percentile along this direction are retained to preserve $70\%$ of the data. This operation is applied independently to source and target clouds, producing controlled partial-overlap pairs for robust training and evaluation.

For all experiments, we follow the official partition of ModelNet40, consistent with prior works such as DBDNet~\cite{li2024dbdnet}, RPMNet~\cite{yew2020rpm}, and REGTR~\cite{yew2022regtr}. While full $k$-fold cross validation is not performed due to the computational cost on large-scale 3D datasets, the combination of standard splits and cross-dataset evaluation provides a practical and widely accepted alternative for fair and robust assessment. To further examine robustness, we also test models trained on ModelNet40 directly on ScanObjectNN and 7Scenes, which differ substantially in object categories, scene complexity, and noise levels.

\subsection{Implementation Details}
To ensure reproducibility, we outline the key implementation settings of our method. The $k$-nearest neighbor search uses $k=12$ across all modules. The attention-based graph neural network extracts 256-dimensional point-wise features. In the HOCE module, we select the top $N = \lfloor 0.5 \times \text{point count} \rfloor$ points from both source and target for overlap estimation, balancing accuracy and efficiency.

In the network architecture, the MLP is implemented using convolutional layers. Furthermore, to facilitate stable and efficient training, Batch Normalization is applied subsequent to each convolutional layer. The entire pipeline is implemented in PyTorch and trained using the RAdam optimizer~\cite{liu2019variance} with an initial learning rate of 0.001. The loss balancing weight $\lambda$ is fixed at 0.5 for all experiments. All experiments are conducted on a workstation equipped with multiple NVIDIA RTX 2080Ti GPUs.

\subsection{Evaluation on ModelNet40}
\label{sec:Registration Performance on ModelNet40}
In this section, we conduct a series of experiments on the ModelNet40 dataset to evaluate the registration performance of our method under various settings. Specifically, we perform evaluations on (1) unknown examples within seen categories to assess intra-class generalization, (2) unknown categories to test cross-category generalization, and (3) Gaussian noise to examine robustness to sensor-level perturbations. We also include qualitative visualizations to illustrate the effectiveness of each stage in our pipeline and to compare visual alignment results with existing methods.

\subsubsection{Baseline Methods and Evaluation Metrics}
Our proposed method is comprehensively evaluated against both traditional and learning-based registration algorithms. Classical baselines include ICP~\cite{besl1992method} and FGR~\cite{zhou2016fast}, while learning-based competitors include DCP~\cite{wang2019deep}, RPMNet~\cite{yew2020rpm}, OMNet~\cite{xu2021omnet}, REGTR~\cite{yew2022regtr}, FINet \cite{xu2022finet}, and DBDNet~\cite{li2024dbdnet}. We use Open3D~\cite{zhou2018open3d} for ICP and FGR and adopt the official implementations for the learning-based methods to ensure reproducibility.

We quantify performance by adopting commonly used evaluation metrics in point cloud registration: root mean squared error (RMSE) and mean absolute error (MAE), computed for both rotation (in degrees) and translation. In addition, an isotropic transformation error (denoted as Error) is reported to reflect the overall alignment quality. Lower values across all metrics indicate better registration performance.

\subsubsection{Evaluation on Unknown Examples}
We evaluated intra-category generalization by training our proposed model on the first 20 categories of ModelNet40 and testing it on previously unseen instances from the same categories. This scenario simulates shape variations within known object classes under partial-overlap conditions. The registration results are presented in \cref{tab:unknown examples}. The best and second-best performances are highlighted in bold and underlined, respectively. Our proposed method achieves the lowest errors across all evaluation metrics, including rotation RMSE (4.298), translation RMSE (0.066), and isotropic alignment error (3.311), outperforming both traditional algorithms and recent learning-based approaches.

\begin{table}[htbp]
	\centering
	\caption{Registration results on unknown examples point clouds.}
	\label{tab:unknown examples}
	\footnotesize
	\begin{tabularx}{\linewidth}{l *{6}{>{\centering\arraybackslash}X}}
	\toprule
	Method & RMSE(R)     & RMSE(t)     & MAE(R)      & MAE(t)      & Error(R)    & Error(t)    \\ \midrule
	ICP~\cite{besl1992method}    & 29.769      & 0.380       & 15.776      & 0.220       & 31.099      & 0.469       \\
	FGR~\cite{zhou2016fast}    & 21.331      & 0.182       & 4.460       & 0.040       & 7.961       & 0.086       \\
	DCP~\cite{wang2019deep}    & 13.650      & 0.218       & 9.693       & 0.157       & 18.895      & 0.321       \\
	RPMNet~\cite{yew2020rpm} & 10.904      & 0.178       & 7.761       & 0.129       & 14.883      & 0.257       \\
	OMNet~\cite{xu2021omnet}  & 8.371       & 0.122       & 3.005       & 0.049       & 6.255       & 0.108       \\
	REGTR~\cite{yew2022regtr}  & 7.824       & 0.092       & 2.998       & 0.041       & 6.021       & 0.092       \\
	FINet~\cite{xu2022finet}  & 6.123       & {\ul 0.078} & 2.329       & {\ul 0.032} & 4.828       & {\ul 0.070} \\
	DBDNet~\cite{li2024dbdnet} & {\ul 5.381} & 0.081       & {\ul 2.027} & 0.037       & {\ul 4.203} & 0.080       \\
	\textbf{CEGC(ours)} & \textbf{4.298} & \textbf{0.066} & \textbf{1.632} & \textbf{0.030} & \textbf{3.311} & \textbf{0.065} \\ \bottomrule
	\end{tabularx}
	\end{table}

Classical methods, such as ICP and FGR, exhibit significantly higher errors, highlighting their reliance on good initial alignment and their inability to handle partial overlaps. Although learning-based methods such as DCP and RPMNet demonstrate improved generalization, their performance remains limited due to insufficient modeling of semantic context and structural coherence.

By contrast, our proposed framework consistently achieves superior accuracy by explicitly addressing these challenges. The HOCE module effectively identifies overlapping regions by fusing semantic and geometric cues, while CAMS leverages global attention to enforce correspondence consistency under shape and viewpoint variations. Combined with confidence-weighted transformation estimation, these components contribute to robust alignment even under severe partiality. This synergy explains the significant performance margin observed in this intra-class generalization setting.

\subsubsection{Evaluation on Unknown Categories}
We conducted experiments on unseen categories to assess cross-category generalization In particular, the 40 categories in the ModelNet40 dataset are evenly split into two disjoint subsets: the first 20 categories are used for training and the remaining 20 are used for testing. This setting examines the robustness of the model when for entirely novel object classes. The experimental results are presented in \cref{tab:unknown categories}.

\begin{table}[htbp]
	\centering
	\caption{Registration results on unknown categories point clouds.}
	\label{tab:unknown categories}
	\footnotesize
	\begin{tabularx}{\linewidth}{l *{6}{>{\centering\arraybackslash}X}}
	\toprule
	Method & RMSE(R)        & RMSE(t)        & MAE(R)         & MAE(t)         & Error(R)       & Error(t)       \\ \midrule
	ICP~\cite{besl1992method}    & 28.464 & 0.375 & 16.333 & 0.222 & 31.543 & 0.466 \\
	FGR~\cite{zhou2016fast}    & 23.916 & 0.214 & 4.936  & 0.045 & 8.101  & 0.096 \\
	DCP~\cite{wang2019deep}    & 16.186 & 0.248 & 11.766 & 0.180 & 22.816 & 0.365 \\
	RPMNet~\cite{yew2020rpm} & 13.073 & 0.207 & 9.535  & 0.151 & 18.522 & 0.306 \\
	OMNet~\cite{xu2021omnet}  & 11.037 & 0.150 & 4.547  & 0.067 & 9.327  & 0.141 \\
	REGTR~\cite{yew2022regtr}  & 9.785  & 0.138 & 4.112  & 0.058 & 8.982  & 0.131 \\
	FINet~\cite{xu2022finet}  & 8.688  & 0.111 & 3.650  & 0.049 & 7.248  & 0.104 \\
	DBDNet~\cite{li2024dbdnet} & {\ul 6.538}    & {\ul 0.090}    & {\ul 2.226}    & {\ul 0.040}    & {\ul 4.541}    & {\ul 0.084}    \\
	\textbf{CEGC(ours)}   & \textbf{5.271} & \textbf{0.074} & \textbf{1.801} & \textbf{0.033} & \textbf{3.693} & \textbf{0.069} \\ \bottomrule
	\end{tabularx}
	\end{table}

As indicated in \cref{tab:unknown categories}, all methods exhibit varying degrees of performance degradation compared to the known-category setting, reflecting the inherent challenge of domain shift. Regardless, our proposed method maintains strong performance and consistently outperforms all baselines. It achieves the lowest RMSE (5.271 for rotation, 0.074 for translation), MAE (1.801 and 0.033), and isotropic error (3.693 and 0.069), outperforming the closest competitor (DBDNet) by a considerable margin.

These results demonstrate the strong category-level generalization of our proposed framework. We attribute this to the synergistic integration of HOCE and context-aware correspondence reasoning, which enable the model to effectively adapt to novel geometric structures while maintaining high accuracy. By contrast, both traditional and existing learning-based approaches often suffer from overfitting to seen categories, resulting in degraded performance in generalization scenarios.

\subsubsection{Evaluation on Gaussian Noise}
We introduced Gaussian noise to both source and target point clouds to evaluate the robustness of our proposed method under noisy conditions. In particular, each point was perturbed by zero-mean Gaussian noise with a standard deviation of 0.01, clipped to the range $[-0.05, 0.05]$ along each coordinate axis. This perturbation degrades point-wise correspondences and simulates the types of noise generally introduced by depth sensors and LiDAR systems in real-world scenarios. Experiments were conducted under both unknown examples and unknown categories settings.

\begin{table}[htbp]
	\centering
	\caption{Registration results on unknown examples point clouds with gaussian noise.}
	\label{tab:unknown examples with gaussian noise}
	\footnotesize
	\begin{tabularx}{\linewidth}{l *{6}{>{\centering\arraybackslash}X}}
	\toprule
	Method & RMSE(R) & RMSE(t) & MAE(R) & MAE(t)      & Error(R) & Error(t) \\ \midrule
	ICP~\cite{besl1992method}    & 29.950  & 0.391   & 15.881 & 0.229       & 32.100   & 0.482    \\
	FGR~\cite{zhou2016fast}    & 37.814  & 0.336   & 13.836 & 0.126       & 24.938   & 0.271    \\
	DCP~\cite{wang2019deep}    & 14.128  & 0.216   & 10.069 & 0.156       & 19.500   & 0.320    \\
	RPMNet~\cite{yew2020rpm} & 11.613  & 0.187   & 8.244  & 0.134       & 16.008   & 0.275    \\
	OMNet~\cite{xu2021omnet}  & 8.949   & 0.137   & 3.535  & 0.053       & 7.204    & 0.114    \\
	REGTR~\cite{yew2022regtr}  & 7.874   & 0.113   & 3.143  & 0.043       & 6.321    & 0.103    \\
	FINet~\cite{xu2022finet}  & 6.997   & 0.086   & 2.681  & {\ul 0.037} & 5.475    & 0.078    \\
	DBDNet~\cite{li2024dbdnet} & {\ul 5.775}    & {\ul 0.083}    & {\ul 2.231}    & 0.040          & {\ul 4.612}    & {\ul 0.085}    \\
	\textbf{CEGC(ours)}   & \textbf{4.799} & \textbf{0.071} & \textbf{1.853} & \textbf{0.034} & \textbf{3.783} & \textbf{0.072} \\ \bottomrule
	\end{tabularx}
	\end{table}

Quantitative results are summarized in \cref{tab:unknown examples with gaussian noise} and \cref{tab:unknown categories with gaussian noise}, corresponding to the unknown shape and unknown category settings, respectively. Our proposed method consistently outperforms classical and learning-based baselines across all evaluation metrics. In particular, under the unseen shape setting (\cref{tab:unknown examples with gaussian noise}), our proposed approach achieves a rotation RMSE of 4.799 and a translation RMSE of 0.071, which correspond to improvements of 16.9\% and 14.5\% over the strongest baseline (DBDNet). The gap widens considerably compared to classical methods such as ICP (29.950 / 0.391) and FGR (37.814 / 0.336), which fail to produce reliable alignments under such perturbations. Similarly, under the unseen category setting (\cref{tab:unknown categories with gaussian noise}), our proposed method also maintains strong generalization, achieving RMSE values of 5.811 and 0.079 for rotation and translation, respectively, once again outperforming all baselines. These results demonstrate the capacity of the model to generalize across different object semantics even when exposed to noisy and previously unseen data.

\begin{table}[htbp]
	\centering
	\caption{Registration results on unknown categories point clouds with gaussian noise.}
	\label{tab:unknown categories with gaussian noise}
	\footnotesize
	\begin{tabularx}{\linewidth}{l *{6}{>{\centering\arraybackslash}X}}
	\toprule
	Method & RMSE(R)        & RMSE(t)        & MAE(R)         & MAE(t)         & Error(R)       & Error(t)       \\ \midrule
	ICP~\cite{besl1992method}    & 27.849 & 0.383 & 16.341 & 0.234 & 32.175 & 0.484 \\
	FGR~\cite{zhou2016fast}    & 38.140 & 0.302 & 12.739 & 0.102 & 20.569 & 0.218 \\
	DCP~\cite{wang2019deep}    & 16.532 & 0.249 & 12.022 & 0.181 & 23.184 & 0.366 \\
	RPMNet~\cite{yew2020rpm} & 13.843 & 0.218 & 10.108 & 0.157 & 19.560 & 0.324 \\
	OMNet~\cite{xu2021omnet}  & 10.500 & 0.163 & 4.661  & 0.078 & 9.472  & 0.162 \\
	REGTR~\cite{yew2022regtr}  & 9.992  & 0.151 & 4.112  & 0.066 & 8.423  & 0.141 \\
	FINet~\cite{xu2022finet}  & 9.001  & 0.120 & 3.827  & 0.055 & 7.748  & 0.117 \\
	DBDNet~\cite{li2024dbdnet} & {\ul 6.934}    & {\ul 0.093}    & {\ul 2.368}    & {\ul 0.042}    & {\ul 4.905}    & {\ul 0.088}    \\
	\textbf{CEGC(ours)}   & \textbf{5.811} & \textbf{0.079} & \textbf{1.983} & \textbf{0.036} & \textbf{4.132} & \textbf{0.075} \\ \bottomrule
	\end{tabularx}
	\end{table}

Our proposed framework achieves robust performance under substantial Gaussian noise by integrating HOCE, context-aware correspondence reasoning, and confidence-guided transformation estimation. The proposed method effectively mitigates noise-induced correspondence errors by jointly modeling semantic and geometric cues, leveraging global attention to enhance structural consistency, and adaptively downweighting unreliable matches. Experiments on both seen and unseen categories confirm strong generalization and resilience, demonstrating the practical applicability of the method in real-world noisy environments.

\subsubsection{Qualitative Analysis}
\label{sec:qualitative}
\begin{figure}[htbp]
	\centering
	\includegraphics[width=0.9\linewidth,keepaspectratio]{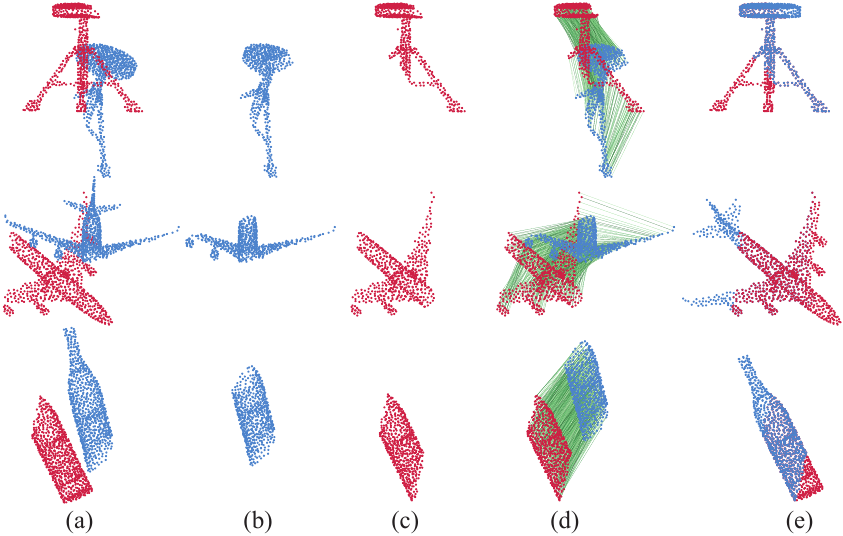}
	\caption{Visualization of partial-to-partial registration on unknown examples. (a) Initial point clouds X and Y; (b) X after overlap estimation; (c) Y after overlap estimation; (d) correspondence search with confidence weights (line opacity); (e) final registration result.}
	\label{fig:RegistrationStep}
	\end{figure}

We intuitive illustrate the effectiveness of our registration pipeline by visualizing each stage of the algorithm in \cref{fig:RegistrationStep} using previously unseen examples instances under partial-overlap conditions. The figure presents a stepwise progression from raw input to final alignment: (a) the initial source and target point clouds X and Y; (b) and (c) output of the HOCE module, which explicitly estimates overlapping regions by jointly leveraging semantic confidence and cross-cloud geometric similarity; (d) the correspondences predicted by the CAMS module, where the opacity of lines encodes the confidence of each match; and (e) the final registration result produced by a confidence-weighted transformation solver.

This staged visualization reflects our core design: the proposed method avoids reliance on brittle global consistency or heuristic pruning by decoupling the registration task into overlap prediction, performing context-enhanced matching, and estimating weighted transformation. It transforms the ill-posed partial-to-partial registration problem into a structured optimization pipeline, where each stage progressively filters noise, resolves ambiguity, and reinforces geometric coherence. The figure exemplifies how our design enables robust, data-driven registration, even in the presence of severe incompleteness and semantic variability.

\begin{figure}[htbp]
	\centering
	\includegraphics[width=\linewidth,keepaspectratio]{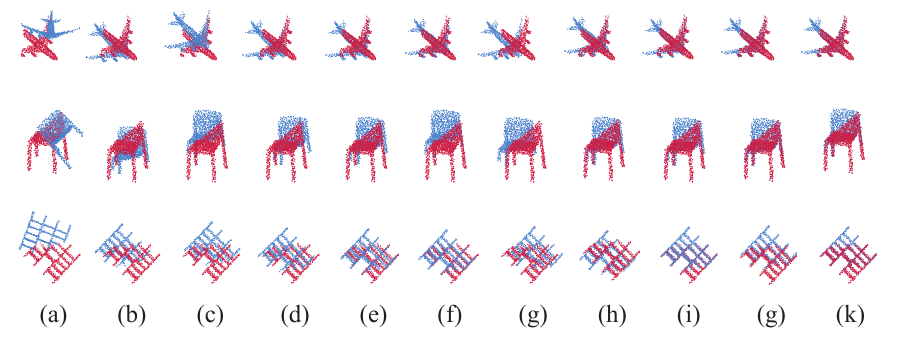}
	\caption{Qualitative comparison of registration methods on unseen examples with Gaussian noise. From (a) to (k): input point clouds, ICP, FGR, DCP, RPMNet, OMNet, REGTR, FINet, DBDNet, our method, and ground-truth alignment.}
	\label{fig:RegistrationDifferent}
	\end{figure}

We qualitatively evaluate robustness under noise by visualizing the registration results of different methods on unseen examples corrupted with Gaussian noise, as illustrated in \cref{fig:RegistrationDifferent}. Subfigure (a) shows the input point clouds, (b)–(j) present the results of ICP, FGR, DCP, RPMNet, OMNet, REGTR, FINet, DBDNet, and our proposed method, respectively, and (k) provides the ground truth.

Classical methods (b and c) fail owing to poor initialization and global-overlap assumptions. Learning-based methods (d–i) improve performance but often suffer from local misalignment under noise. By contrast, our proposed method (j) achieves results that closely match the ground truth (k), demonstrating higher resilience. The results highlight the effectiveness of our design in addressing partial, noisy data. The proposed method achieves stable and accurate alignment across diverse shapes by integrating overlap-aware region filtering, context-driven correspondence reasoning, and confidence-weighted transformation. This demonstrates strong generalization and robustness, crucial for real-world 3D perception tasks under imperfect sensing conditions.

\subsection{Generalization Across Different Datasets}
We adopted a cross-dataset generalization setting to rigorously evaluate the applicability of our method in real-world scenarios—where sensor noise, occlusion, and environmental variability are prevalent. In particular, the model is trained only on the unknown examples split of ModelNet40, following the standard protocol in previous studies, and then directly applied to two challenging benchmarks without any fine-tuning. The benchmarks include ScanObjectNN~\cite{uy2019revisiting}, which introduces background clutter and occlusion at the object level, and 7Scenes~\cite{shotton2013scene}, which captures noisy indoor scene-level data from RGB-D cameras. This setup enables a stringent test of generalization under substantial domain shifts. The consistent performance across both synthetic and real-world datasets provides strong evidence that our proposed method remains robust and transferable beyond the training distribution.

\subsubsection{Evaluation on ScanObjectNN Dataset}
To simulate object-level deployment in realistic environments, we first evaluated our method on the ScanObjectNN dataset. Unlike the clean, synthetic CAD models of ModelNet40, ScanObjectNN contains 3D point clouds of real-world objects captured by commodity RGB-D sensors. These data are inherently noisy and feature common real-world artifacts, including background clutter, occlusions, and non-uniform point sampling—factors that severely challenge traditional 3D registration pipelines. Importantly, both ModelNet40 and ScanObjectNN follow a partial-overlap protocol, ensuring that performance differences are attributable to domain shift rather than methodological inconsistencies. Here, we benchmark only against learning-based methods.

\begin{table}[htbp]
	\centering
	\caption{Registration results on ScanObjectNN dataset.}
	\label{tab:ScanObjectNN dataset}
	\footnotesize
	\begin{tabularx}{\linewidth}{l *{6}{>{\centering\arraybackslash}X}}
	\toprule
	Method & RMSE(R)        & RMSE(t)        & MAE(R)         & MAE(t)         & Error(R)       & Error(t)       \\ \midrule
	DCP~\cite{wang2019deep}                  & 19.581 & 0.266 & 13.797 & 0.200 & 31.825 & 0.423 \\
	RPMNet~\cite{yew2020rpm}               & 15.640 & 0.235 & 11.047 & 0.164 & 21.317 & 0.334 \\
	OMNet~\cite{xu2021omnet}                & 14.519 & 0.149 & 4.938  & 0.060 & 10.123 & 0.140 \\
	REGTR~\cite{yew2022regtr}                & 12.232 & 0.119 & 4.347  & 0.050 & 8.767  & 0.112 \\
	FINet~\cite{xu2022finet}                & 14.961 & 0.141 & 5.872  & 0.055 & 11.314 & 0.126 \\
	DBDNet~\cite{li2024dbdnet} & {\ul 8.305}    & {\ul 0.109}    & {\ul 2.903}    & {\ul 0.049}    & {\ul 6.029}    & {\ul 0.097}    \\
	\textbf{CEGC(ours)}   & \textbf{5.778} & \textbf{0.081} & \textbf{2.192} & \textbf{0.037} & \textbf{4.448} & \textbf{0.077} \\ \bottomrule
	\end{tabularx}
	\end{table}

The registration results are summarized in \cref{tab:ScanObjectNN dataset}. Our proposed method consistently outperforms all baselines across all metrics, achieving the lowest rotation RMSE (5.778), translation RMSE (0.081), and isotropic errors (4.448 and 0.077). Compared with its performance on the ModelNet40 unknown examples setting (4.298, 0.066), the proposed model shows only a modest performance drop. By contrast, other methods exhibit a more substantial degradation. For instance, the rotation error of DCP increases from 18.9 to 31.8, and that of OMNet increases from 6.255 to 10.123.

This strong generalization performance highlights the robustness of our proposed framework. By learning to infer reliable overlapping regions and leveraging contextual relationships for feature matching, our proposed model demonstrates enhanced adaptability in the presence of real-world noise and variation. These results underscore its practical value in object-level 3D understanding applications.

\subsubsection{Evaluation on 7scenes Dataset}
We conducted evaluation of the 7Scenes dataset to further validate the practical utility of our method in dense, cluttered indoor scenes. This benchmark comprises RGB-D sequences recorded by Kinect in everyday indoor environments and introduces a more severe domain shift: the scenes are highly unstructured, affected by significant occlusion, dynamic lighting, sensor noise, and large-scale layout variability. These characteristics make 7Scenes a compelling proxy for realistic indoor SLAM, localization, and scene reconstruction tasks. The dataset was processed by following the procedure described in a previous study~\cite{shen2022reliable} to ensure consistency with prior work and reproducibility of evaluation.

\begin{table}[htbp]
	\centering
	\caption{Registration results on 7scenes dataset.}
	\label{tab:7scenes dataset}
	\footnotesize
	\begin{tabularx}{\linewidth}{l *{6}{>{\centering\arraybackslash}X}}
	\toprule
	Method & RMSE(R) & RMSE(t) & MAE(R) & MAE(t) & Error(R) & Error(t) \\ \midrule
	DCP~\cite{wang2019deep}    & 22.433  & 0.399   & 16.791 & 0.288  & 32.130   & 0.584    \\
	RPMNet~\cite{yew2020rpm} & 25.232  & 0.365   & 16.866 & 0.236  & 30.413   & 0.457    \\
	OMNet~\cite{xu2021omnet}  & 26.067  & 0.327   & 9.536  & 0.098  & 16.891   & 0.235    \\
	REGTR~\cite{yew2022regtr}  & 15.682  & 0.169   & 6.348  & 0.073  & 12.231   & 0.159    \\
	FINet~\cite{xu2022finet}  & 14.698  & 0.252   & 5.630  & 0.090  & 12.569   & 0.193    \\
	DBDNet~\cite{li2024dbdnet} & {\ul 8.095}    & {\ul 0.108}    & {\ul 3.349}    & {\ul 0.049}    & {\ul 6.404}    & {\ul 0.112}    \\
	\textbf{CEGC(ours)}   & \textbf{6.260} & \textbf{0.081} & \textbf{2.404} & \textbf{0.036} & \textbf{4.634} & \textbf{0.083} \\ \bottomrule
	\end{tabularx}
	\end{table}

Our model, trained on ModelNet40 without exposure to any scene-level data, was directly applied to 7Scenes for registration. As indicated in \cref{tab:7scenes dataset}, it achieves superior performance across all metrics, with the lowest rotation RMSE and translation RMSE, as well as minimal mean and isotropic errors. Compared to its performance on the unknown examples setting, the degradation remains marginal. More specifically, the rotation error increases from 4.298 to 6.260, while translation error rises from 0.066 to 0.081. By contrast, competitors show significantly greater drops. For example, the rotation error of OMNet increases from 6.255 to 16.891, and that of REGTR increases from 6.021 to 12.231.

These findings demonstrate that our proposed method generalizes effectively to scene-level data, despite being trained exclusively on synthetic object-level inputs. The success stems from the ability to adaptively suppress unreliable regions and maintain consistent correspondences, even under complex scene layout and occlusion. These results affirm the practical applicability of the model to real-world 3D perception tasks, including robotics, augmented reality, and indoor mapping.

\subsubsection{Qualitative Visualization of Cross-Dataset Generalization}
\begin{figure}[!ht]
    \centering
    \includegraphics[width=\linewidth,keepaspectratio]{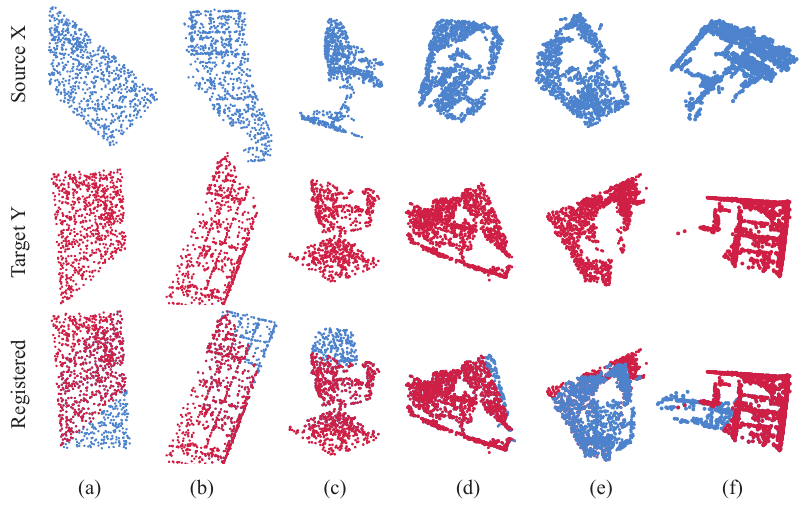}
    \caption{Qualitative visualization of cross-dataset registration. (a-c): results on ScanObjectNN; (d-f): results on 7Scenes. Each column shows the source point cloud (top), the target point cloud (middle), and the registration result (bottom). Our method achieves precise alignment despite strong domain variations and structural incompleteness.}
    \label{fig:Fig_Generalization}
\end{figure}

To visually demonstrate the cross-domain robustness of our method, we present qualitative results on the ScanObjectNN and 7Scenes datasets in \cref{fig:Fig_Generalization}. Subfigures (a)-(c) correspond to a representative sample from the ScanObjectNN dataset, while (d)-(f) show results from the 7Scenes dataset. Despite the substantial domain differences between synthetic training data and real-world test scenarios—including occlusion, sensor noise, and structural variability—our method achieves accurate alignment in both object-level and scene-level settings. The registered outputs show high geometric consistency and tight alignment, highlighting strong generalization capability.

\subsubsection{Discussion on Dataset Differences}
The performance gap across different datasets can be largely explained by their inherent characteristics. ScanObjectNN, although featuring real-world noise and background clutter, consists mainly of isolated objects. In this case, our HOCE module can effectively suppress irrelevant regions, leading to only a modest performance drop compared to the synthetic ModelNet40 dataset. In contrast, 7Scenes poses a more severe challenge: the data are captured in cluttered indoor environments with significant occlusion, dynamic lighting, and large-scale layout variability. These factors introduce strong structural ambiguity and explain the relatively higher error margins on 7Scenes. Nevertheless, our method still achieves the best performance among all baselines, underscoring its robustness across diverse domains. This analysis suggests that dataset-specific characteristics, particularly occlusion severity and scene complexity, play a critical role in registration difficulty, providing further insights into the strengths and limitations of our framework.

\subsection{Ablation Studies}
\label{sec:Ablation Study}
\subsubsection{Ablation on Core Modules}
We performed ablation experiments on the unseen subset of ModelNet40 under Gaussian noise to evaluate the effectiveness of the two core components in our framework, namely, HOCE and CAMS. As described in \cref{sec:METHOD}, the HOCE module identifies overlapping regions by jointly modeling semantic descriptors and geometric similarities. By contrast, the CAMS module enhances correspondence search by applying attention-based global reasoning and assigning soft confidence weights to candidate matches. These two modules target complementary stages of the registration pipeline: identifying overlapping regions and refining point correspondences.

We assessed their individual and joint contributions by designing the following four experimental configurations:

(A) Baseline: Both HOCE and CAMS were removed. The model used a basic overlap predictor similar to OMNet~\cite{xu2021omnet} and assumed uniform confidence for all correspondences without any form of contextual modeling.

(B) Baseline + HOCE: The HOCE module was included for overlap region estimation; correspondence weighting remained uniform.

(C) Baseline + CAMS: The CAMS module was included for context-aware correspondence weighting; a basic overlap predictor was used.

 (D) Full model: Both HOCE and CAMS were integrated.

\begin{table}[htbp]
	\centering
	\caption{Ablation study of HOCE and CAMS on ModelNet40 under nknown examples with Gaussian noise.}
	\label{tab:ablation}
	\footnotesize
	\begin{tabularx}{\linewidth}{l *{6}{>{\centering\arraybackslash}X}}
	\toprule
	ID & HOCE & CAMS & RMSE(R) & RMSE(t) & MAE(R) & MAE(t) \\
	\midrule
	A & \ding{55} & \ding{55} & 7.182 & 0.146 & 3.183 & 0.067 \\
	B & \ding{51} & \ding{55} & {\ul 5.878} & 0.096 & {\ul 2.071} & 0.045 \\
	C & \ding{55} & \ding{51} & 5.994 & {\ul 0.091} & 2.215 & {\ul 0.042} \\
	D & \ding{51} & \ding{51} & \textbf{4.799} & \textbf{0.071} & \textbf{1.853} & \textbf{0.034} \\
	\bottomrule
	\end{tabularx}
\end{table}

As indicated in \cref{tab:ablation}, each proposed module contributes significantly to the registration pipeline in a manner that directly reflects its functional role.

(B vs. A) HOCE improves both rotation and translation accuracies by identifying overlapping regions between partially visible point clouds. It combines semantic descriptors with geometric similarities to suppress noisy or non-overlapping areas and reduce false correspondences. The effect is most evident in the rotation accuracy (RMSE drops from 7.182 to 5.878).

(C vs. A) CAMS complements this by assigning confidence weights to matches through global contextual reasoning. This reduces errors from ambiguous or repetitive local structures, yielding notable gains in translation accuracy (MAE(t) decreases from 0.067 to 0.042).

(D) When combined, the system achieves the best overall performance. This is because the two modules address different stages and challenges in the registration process: HOCE enhances the quality of the candidate set by filtering out irrelevant or misleading points before matching, while CAMS makes the matching process more reliable by reinforcing globally consistent associations. Their synergy constructs a robust correspondence map grounded in both spatial relevance and contextual coherence. This enables accurate rigid transformation under noise and partial visibility.

These results validate the effectiveness of HOCE and CAMS as well as demonstrate that explicitly addressing region selection and match confidence yields a principled solution for robust partial point cloud registration.

\subsubsection{Ablation on HOCE Sub-branches.}

We further quantified the internal design of HOCE by independently evaluating its two sub-branches under the same unseen examples with Gaussian noise setting. As indicated in \cref{tab:ablation_HOCE}, enabling only the Semantic branch (B) reduces rotational errors compared to the baseline (A), while enabling only the Geometric branch (C) yields larger gains in translation metrics. The fused branches (D) consistently outperform the single-branch variants, confirming their complementarity: semantic cues suppress semantically inconsistent regions, whereas geometric similarity stabilizes low-level structural consistency. When combined, they deliver the best overlap confidence and downstream alignment.

\begin{table}[htbp]
  \centering
  \caption{Ablation on HOCE sub-branches.}
  \label{tab:ablation_HOCE}
  \footnotesize
  \begin{tabularx}{\linewidth}{l *{7}{>{\centering\arraybackslash}X}}
  \toprule
  ID & Semantic & Geometric & CAMS & RMSE(R) & RMSE(t) & MAE(R) & MAE(t) \\
  \midrule
  A & \ding{55} & \ding{55} & \ding{51} & 5.878 & 0.091 & 2.071 & 0.042 \\
  B & \ding{51} & \ding{55} & \ding{51} & {\ul 5.363} & 0.089 & 2.056 & {\ul 0.039} \\
  C & \ding{55} & \ding{51} & \ding{51} & 5.512 & {\ul 0.084} & {\ul 1.973} & 0.041 \\
  D & \ding{51} & \ding{51} & \ding{51} & \textbf{4.799} & \textbf{0.071} & \textbf{1.853} & \textbf{0.034} \\
  \bottomrule
  \end{tabularx}
\end{table}

\subsection{Discussion}
\label{sec:Discussion}
\subsubsection{Evaluation under Extreme Overlap Conditions}
\label{sec:low_overlap}
We conducted an additional experiment focusing on very low overlap ratios to further strengthen the evaluation under extreme scenarios. We tested our proposed method when the overlap ratio between source and target point clouds was reduced to only 40\%, which is significantly lower than the standard 70\% partial-overlap setting. All experiments were performed on the unseen examples subset of ModelNet40 with Gaussian noise, ensuring consistency with the protocol used in our main experiments.

\begin{table}[htbp]
	\centering
	\caption{Registration results under a 40\% overlap ratio on unseen examples with Gaussian noise.}
	\label{tab:low_overlap}
	\footnotesize
	\begin{tabularx}{\linewidth}{l *{6}{>{\centering\arraybackslash}X}}
	\toprule
	Method & RMSE(R) & RMSE(t) & MAE(R) & MAE(t)      & Error(R) & Error(t) \\ \midrule
	ICP~\cite{besl1992method}    & 58.384  & 0.733   & 28.461 & 0.403       & 57.281   & 0.855    \\
	FGR~\cite{zhou2016fast}    & 44.249  & 0.443   & 16.783 & 0.165       & 29.452   & 0.308    \\
	DCP~\cite{wang2019deep}    & 19.355  & 0.378   & 13.714 & 0.263       & 23.616   & 0.497    \\
	RPMNet~\cite{yew2020rpm} & 38.862  & 0.625   & 20.491 & 0.410       & 38.214   & 0.705    \\
	OMNet~\cite{xu2021omnet}  & 17.781  & 0.319   & 6.283  & 0.105       & 16.490   & 0.253    \\
	REGTR~\cite{yew2022regtr}  & 14.673  & 0.277   & 5.417  & 0.081       & 11.439   & 0.181    \\
	FINet~\cite{xu2022finet}  & 9.843   & 0.145   & 4.252  & {\ul 0.062} & 7.491    & {\ul 0.103}    \\
	DBDNet~\cite{li2024dbdnet} & {\ul 9.167}    & {\ul 0.113}    & {\ul 3.501}    & 0.067          & {\ul 6.508}    & 0.127    \\
	\textbf{CEGC(ours)}   & \textbf{6.478} & \textbf{0.089} & \textbf{2.487} & \textbf{0.042} & \textbf{4.926} & \textbf{0.084} \\ \bottomrule
	\end{tabularx}
\end{table}

As indicated in \cref{tab:low_overlap}, our proposed method consistently outperforms all competing approaches under the 40\% overlap setting. Compared with \cref{tab:unknown examples with gaussian noise} (70\% overlap), all methods exhibit performance degradation as expected. Nevertheless, our proposed model achieves the lowest rotation and translation errors across all metrics. More importantly, the relative increase in error for our proposed method is smaller than that of the baselines, which highlights the robustness of our approach to extreme geometric incompleteness and partial visibility. These results demonstrate that our proposed framework remains reliable even under challenging real-world conditions with severely limited overlap.

\subsubsection{Computational Efficiency.}
We further analyzed the computational efficiency of our proposed method in comparison with recent learning-based registration approaches. All experiments were conducted on an NVIDIA RTX 2080Ti GPU, and for fairness, model loading and preprocessing time were excluded. The reported values correspond to the average runtime across the entire test set, decomposed into model inference time and pose estimation time.

\begin{table}[htbp]
	\centering
	\caption{Average runtime (in seconds) across the ModelNet40 test set. The model time is the time for feature extraction, while the pose time is the time for transformation estimation.}
	\label{tab:runtime}
	\footnotesize
	\begin{tabularx}{\linewidth}{l *{3}{>{\centering\arraybackslash}X}}
		\toprule
		Method & Model Time & Pose Time & Total Time \\ \midrule
		DCP~\cite{wang2019deep}    & {\ul 0.0047} & \textbf{0.0039} & \textbf{0.0085} \\
		RPMNet~\cite{yew2020rpm} & 0.0524 & 0.0311 & 0.0835 \\
		OMNet~\cite{xu2021omnet}  & 0.0197 & 0.0155 & 0.0353 \\
		REGTR~\cite{yew2022regtr}  & 0.0621 & 0.0367 & 0.0989 \\
		FINet~\cite{xu2022finet}  & \textbf{0.0037} & {\ul 0.0081} & {\ul0.0118} \\
		DBDNet~\cite{li2024dbdnet} & 0.0664 & 0.0139 & 0.0803 \\
		\textbf{CEGC(ours)}   & 0.0468 & 0.0215 & 0.0682 \\
		\bottomrule
	\end{tabularx}
\end{table}

As summarized in \cref{tab:runtime}, our proposed method requires an average of 0.0682 s per pair, which is comparable to RPMNet (0.0835 s) and DBDNet (0.0803 s), and faster than RegTR (0.0989 s).
Although methods such as DCP (0.0085 s) and FINet (0.0118 s) achieve faster runtimes owing to their lightweight one-pass architectures, they are generally less robust under partial-overlap and noisy settings. By explicitly modeling semantic-geometric confidence, our proposed method offers a competitive balance between efficiency and robustness, highlighting its potential for practical deployment in time-critical 3D perception scenarios.

\section{CONCLUSION}
\label{sec:CONCLUSION}
We proposed a unified framework for robust partial-to-partial point cloud registration by explicitly modeling overlap confidence and global contextual correspondence reliability. Our proposed method decomposes the task into three structured stages: hybrid overlap confidence estimation, context-aware matching, and confidence-guided transformation. Collectively, they enhance robustness against noise, occlusion, and semantic variation. By jointly leveraging semantic descriptors and geometric consistency, the proposed framework enables precise identification of overlapping regions and suppresses unreliable correspondences under structural ambiguity. Experiments on both synthetic and real-world datasets validated the superiority of our proposed approach over existing methods in terms of accuracy and generalization. In addition, ablation studies revealed that each module contributes distinctly to the overall robustness and accuracy. The flexibility and modularity of our proposed framework offers promising potential for downstream tasks such as robotic perception, 3D reconstruction, and cross-domain object understanding.

Future work will explore extending the model to dynamic or non-rigid scenarios and integrating self-supervised learning strategies to improve the scalability and reduce data dependency in real-world applications. Furthermore, we plan on adapting the framework to multiview and temporal settings, facilitating its application to dynamic scene understanding and long-term robotic perception.


\section*{ACKNOWLEDGMENT}
This work was supported by the National Natural Science Foundation of China (No. 51774219), the Science and Technology Innovation Talents Program of Hubei Province, China (No. 2024DJC077), the Key Research and Development Project of Wuhan City, China (No. 2025050102030008), and the Natural Science Foundation of Wuhan City, China (No. 2024040701010055).
Numerical calculations were supported by High-Performance Computing Center of Wuhan University of Science and Technology.



\end{document}